\documentclass[final]{cvpr}

\usepackage{times}
\usepackage{epsfig}
\usepackage{graphicx}
\usepackage{amsmath}
\usepackage{amssymb}
\usepackage{booktabs}
\usepackage{multirow}
\usepackage{algorithm}
\usepackage{algorithmic}

\usepackage[pagebackref=true,breaklinks=true,colorlinks,bookmarks=false]{hyperref}

\newcommand{\x}{\boldsymbol{x}}

\newcommand{\p}{\boldsymbol{p}}

\newcommand{\Lm}{\mathcal{L}}

\def\cvprPaperID{11470} 
\def\confYear{CVPR 2021}

\begin{document}

\title{Dynamic Domain Adaptation for Efficient Inference}

\author{
 Shuang Li\textsuperscript{1} \space\space\space
 JinMing Zhang\textsuperscript{1} \space
 Wenxuan Ma\textsuperscript{1} \space
 Chi Harold Liu\textsuperscript{1}\footnotemark[2] \space\space\space
 Wei Li\textsuperscript{2} \vspace{.3em}\\
 \textsuperscript{1}Beijing Institute of Technology \quad \textsuperscript{2}Inceptio Tech.\\
 \vspace{-.3em}
 {\tt\small \{shuangli, jm-zhang, wenxuanma\}@bit.edu.cn} \space {\tt\small liuchi02@gmail.com} \space {\tt\small liweimcc@gmail.com} 
 \vspace{-.5em}
}

\maketitle

\pagestyle{empty}  
\thispagestyle{empty} 

\renewcommand*{\thefootnote}{\fnsymbol{footnote}}
\setcounter{footnote}{2}
\footnotetext{C. Liu is the corresponding author.}

\maketitle

\begin{abstract}
    Domain adaptation (DA) enables knowledge transfer from a labeled source domain to an unlabeled target domain by reducing the cross-domain distribution discrepancy. Most prior DA approaches leverage complicated and powerful deep neural networks to improve the adaptation capacity and have shown remarkable success. However, they may have a lack of applicability to real-world situations such as real-time interaction, where low target inference latency is an essential requirement under limited computational budget. In this paper, we tackle the problem by proposing a \textit{dynamic domain adaptation} (DDA) framework, which can simultaneously achieve efficient target inference in low-resource scenarios and inherit the favorable cross-domain generalization brought by DA. In contrast to static models, as a simple yet generic method, DDA can integrate various domain confusion constraints into any typical adaptive network, where multiple intermediate classifiers can be equipped to infer ``easier'' and ``harder'' target data dynamically. Moreover, we present two novel strategies to further boost the adaptation performance of multiple prediction exits: 1) a confidence score 
   learning strategy to derive accurate target pseudo labels by fully exploring the prediction consistency of different classifiers; 2) a 
   class-balanced self-training strategy to explicitly adapt multi-stage classifiers from source to target without losing prediction diversity. Extensive experiments on multiple benchmarks are conducted to verify that DDA can consistently improve the adaptation performance and accelerate target inference under domain shift and limited resources scenarios.
    
\end{abstract}

\section{Introduction}\label{sec:introduction}
Many intelligent technologies are boosted by the rapid development of computational capacity~\cite{gpu2004,cpu2011,gpu2014} and deep neural networks~\cite{vgg, alexnet, resnet, densenet}. To ensure high reliability, their loaded deep models have to be trained with massive amount of data, so as to enumerate all possible practical scenarios. Unfortunately, there is always a future situation that is unpredictable, and even an object in the same environment may display visual diversity at different times. For instance, the photos captured by cameras of self-driving car may exhibit large variations under different lighting conditions of day and night. This would inevitably lead to degraded recognition, since test data (\textit{target domain}) and training data (\textit{source domain}) follow different distributions.

\begin{figure}[tb]
    \centering
    \includegraphics[width=0.48\textwidth]{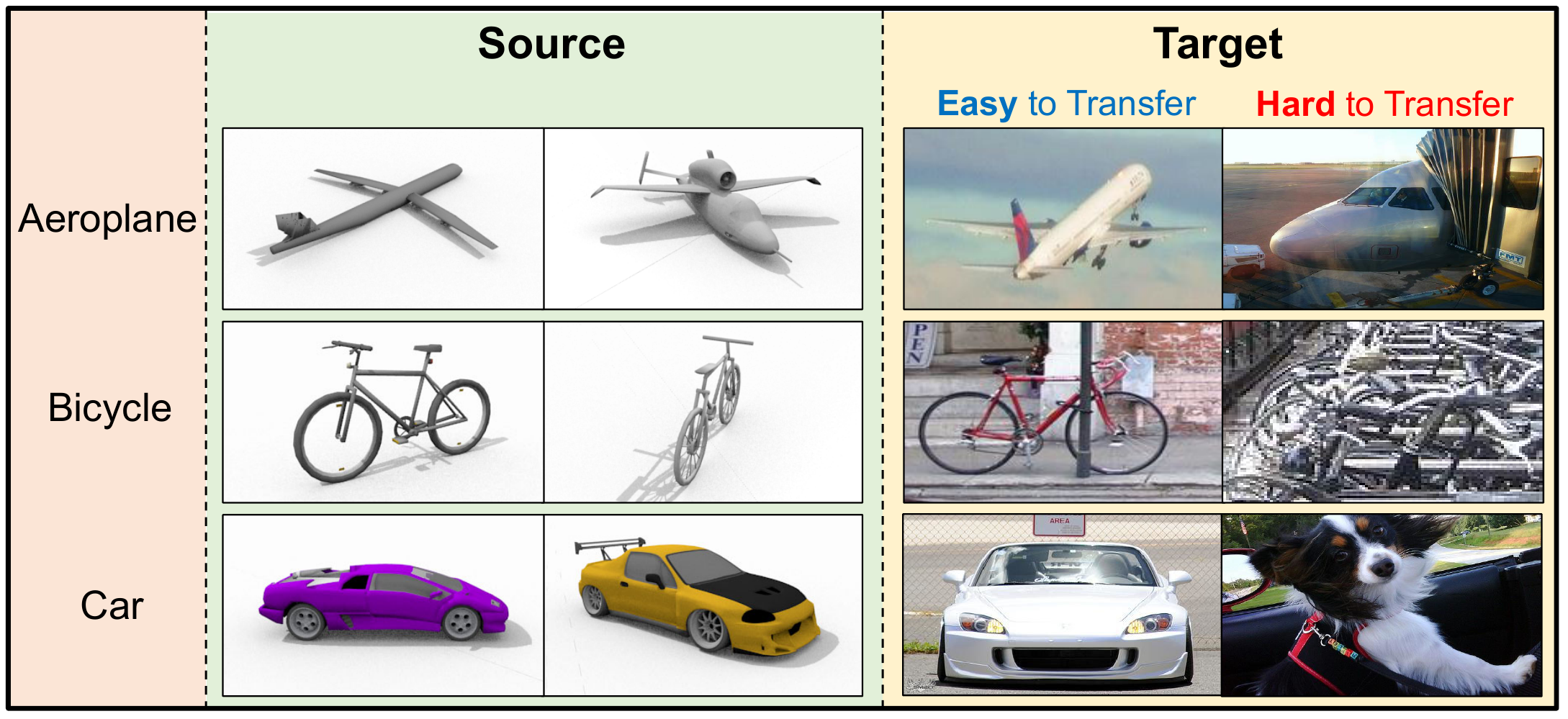}
    \caption{Motivation of the proposed dynamic domain adaptation, which seeks to balance the cross-domain classification performance and the computational cost for target inference. The target images in left column are easy to transfer with a small model, whereas the ``harder'' images require computational expensive models to be correctly recognized.}
    \label{Fig_illstration}
\end{figure}

Such a phenomenon is known as domain shift~\cite{domainshift}, which can be tackled by domain adaptation (DA) techniques~\cite{survey}.
To date there have been considerable research efforts in DA, flourishing with impressive results, especially when applying deep neural networks~\cite{DA_bp,CDAN,MCD,DRCN,DCAN}. They delve into searching for a feature space in which labeled and rich information in source domain can be transferred to unlabeled but related target domain. Generally, these prevailing deep DA methods leverage static and high-complexity base learners owing to their good transferable capacity brought by deep and wide architectures. However, they do not consider the transferability of different target samples as shown in Fig.~\ref{Fig_illstration}. In consequence, they may not be applicable in some real-world situations that require real-time responses or are delay-sensitive to stringent computational resource constraints at inference time.

To allow deep networks to get a grip back on fast inference, there are several techniques that can effectively reduce redundant computational burdens, including network pruning~\cite{Li_pruning,Liu_pruning,Yu_pruning}, architecture design~\cite{mobilenets,squeezenet,shufflenet2}, and knowledge distillation~\cite{knowledgedistill,knowledgedistill1}. Although computation acceleration can be achieved, they are vulnerable to lightweight networks~\cite{mobilenets,condensenet} that are highly optimized. In contrast, another line of work is to explore adaptive inference~\cite{blockdrop,hydranets}, which focuses on dynamically determining the inference structures conditioned on the complexity of input samples and has gained increasing attention recently. Nevertheless, all these methods suffer from poor generalization performance to a new domain, especially when the domain discrepancy is large. Even when state-of-the-art DA methods 
are applied to these models, as shown in the experiments, they still cannot achieve satisfying adaptation performance with efficient inference guaranteed.


Therefore, there is a strong motivation to apply model on resource-constrained device to handle domain discrepancy without losing accuracy.  To tackle this problem, in this paper, we propose a novel framework named \textit{Dynamic Domain Adaptation} (DDA), which can effectively equip vanilla domain adaptation with efficient target inference to balance transferable performance and computational cost in the test phrase. Here, we take the representative adaptive network MSDNet~\cite{msdnet} as our backbone network that has multiple intermediate classifiers at different depth of the network. It could save a large amount of computational cost on ``easy'' samples. Further, we expect that a qualified solution should be feasible in situations of anytime prediction and budget prediction~\footnote{Anytime and budget predictions are two classical settings to evaluate the effectiveness of adaptive inference models, which has been described in detail in~\cite{msdnet}.} under vanilla DA scenarios. 


To be specific, on top of the multi-exist adaptive architecture, we first seek to apply domain confusion constraints to each of the classifier to reduce cross-domain distribution discrepancy of multi-scale features. Based on the direct feature alignment, the multiple classifiers should be able to achieve consistent predictions on samples that are ``easy'' to transfer. Thereby, these target data could be leveraged as ``labeled'' data to further retrain the network with pseudo target supervision, which could significantly improve the target prediction performance. Notably, different from augmenting labeled target set relying on a single classifier, here we exploit probability predictions from multiple classifiers and propose a novel and effective confidence score strategy to discover highly confident pseudo-labeled target samples. By leveraging the calculated confidence score, a trustworthy target set with pseudo labels can be generated, and as the training proceeds, this target set will be more and more precise.

Based on the trustworthy target set, we then utilize the proposed class-balanced self-training strategy to retrain all the classifiers progressively while preserving the prediction diversity among exits. As a result, the classifiers at different stages will be gradually adapted from source to target by the class-balanced self-training. In such a way, our method does not only maintain the
efficiency of adaptive network, but also significantly improve the transferability of each classifier.
In general, we highlight the three-fold contributions.

\begin{itemize}
    \item We propose a dynamic domain adaptation framework to simultaneously achieve satisfying DA performance and fast target inference with low computational cost, which successfully sheds new light on future direction for efficient inference of DA towards resource-limited devices.
    \item Two simple yet effective strategies, confidence score learning and class-balanced self-training, are introduced. By utilizing them, highly confident pseudo-labeled target samples can be selected to retrain all the classifiers, which could significantly improve their adaptation performance.
    \item Comprehensive experimental results verify that the proposed method could greatly save time and computational resources at both anytime and budget prediction settings with promising cross-domain recognition accuracy.
\end{itemize}
\section{Related Work}
\textbf{Adaptive Computation for Deep Network.}
Adaptive computation aims to make unwieldy model lighter to meet the requirements of limited-resource scenarios. Existing works can be typically classified into two threads: static methods and dynamic methods.
For static methods, their goal is to remove redundant network parameters via pruning~\cite{Li_pruning,Liu_pruning,Yu_pruning}, weight quantization~\cite{bnn,quantization2018,Xnornet} or lightweight architecture design~\cite{mobilenets,condensenet}. Although these methods could greatly reduce computational complexity, they relinquish powerful deep network by replacing it with a smaller one or eliminating large amounts of parameters, resulting in their limited representation ability. This motivates a series of works towards dynamic architecture design to obtain a better balance between speed and accuracy~\cite{adaptivegraph,runtime2017,spottune,blockdrop,branchynet,msdnet}. Specifically, the adaptive network intends to allocate appropriate resources to different samples according to their complexity, and classify ``easy'' and ``hard'' samples correctly with dynamic network architectures.

However, these methods will inevitably confront with performance drop caused by the domain shift. In contrast, our DDA framework is proposed to improve their transferability while maintaining the merits of efficient inference.

\textbf{Domain Adaptation.} 
Domain adaptation (DA)~\cite{survey} seeks to learn a well performing model that can generalize from labeled source domain to unlabeled target domain. Prior works mainly rely on distribution alignment by moment matching~\cite{DAN,DCORAL,MDD} or adversarial techniques~\cite{DANN,MCD} to reduce domain shift. 
To name a few, DAN~\cite{DAN}, JAN~\cite{JAN} and DRCN~\cite{DRCN} utilize multi-kernel or joint maximum mean discrepancy~\cite{MMD} to transfer knowledge in task-specific layers. MDD~\cite{MDD} introduces a margin disparity discrepancy to couple two domains with a new generalization bound. However, these methods may suffer from a heavy calculation as the number of samples increases. In contrast, adversarial based DA methods aim to capture domain-invariant representations via a min-max game between feature extractor and domain classifier. For instance, DANN~\cite{DA_bp} enables adaptation behavior by the proposed gradient reversal layer. Afterwards, \cite{MADA,JADA,GVB} introduce different domain confusion terms by adding additional discriminators or classifiers. 


Besides the aforementioned mainstream DA works, there are several methods combined with self-training strategy to adapt source classifier to target via various target selection techniques~\cite{classbalance,DIRT-T,SPCAN,PFA,CRST}. However, these approaches are proposed for static network architecture with unique exit, which cannot be directly used on cascade of intermediate classifiers, and also cannot reduce resource cost at inference. To remedy this, we develop simple yet effective target selection and retraining strategies, which are specially designed for the adaptive network to accelerate target inference with good adaptability guaranteed.

A closely related work targeting at efficient inference in DA, called REDA~\cite{REDA}, adopts a similar MSDNet architecture compared with our method. It utilizes knowledge distillation~\cite{knowledgedistill} method to enhance the performance of shallower classifiers, while performing vanilla DA at the last classifier to guarantee transferability. However, this method limits the improvement of the last classifier, since it doesn't obtain any knowledge from the shallower classifiers. Our proposed strategies, on the contrary, explore the prediction consistency between classifiers of different depth and find within the common knowledge to teach all of them. In this case, all classifiers in DDA mutually promote each other, and thus it is able to achieve an overall performance improvement and thus find a better balance between adaptation performance and computational cost for fast target inference.




\section{Dynamic Domain Adaptation}


\subsection{Preliminaries and Motivation}

In domain adaptation (DA), we usually have access to a labeled source domain $D_s=\{\left(\x_i^s,y_i^s\right)\}_{i=1}^{N_s}$ with $N_s$ labeled samples from $C$ classes, while working on an unlabeled target domain $D_t=\{\x_j^t\}_{j=1}^{N_t}$ of $N_t$ unlabeled samples. The source and target samples are drawn from different distributions $P_s$ and $P_t$. Given the fact that $P_s\neq P_t$, the goal of DA is to train a deep neural network that generalizes well on the target domain by reducing the domain discrepancy.

Due to the static network architecture, though the learned model is with high transferability, vanilla deep DA methods cannot accelerate the target inference time. These methods are limited to real-world applications on resource-constrained platforms such as smart phones or wearable devices. Consequently, it is essential to equip DA with fast inference capacity via adaptive inference models. In this paper, the representative adaptive network MSDNet is used as our backbone network, and we note that the proposed DDA is orthogonal to other adaptive inference models.

\begin{figure}[tb]
  \centering
  \includegraphics[width=0.48\textwidth]{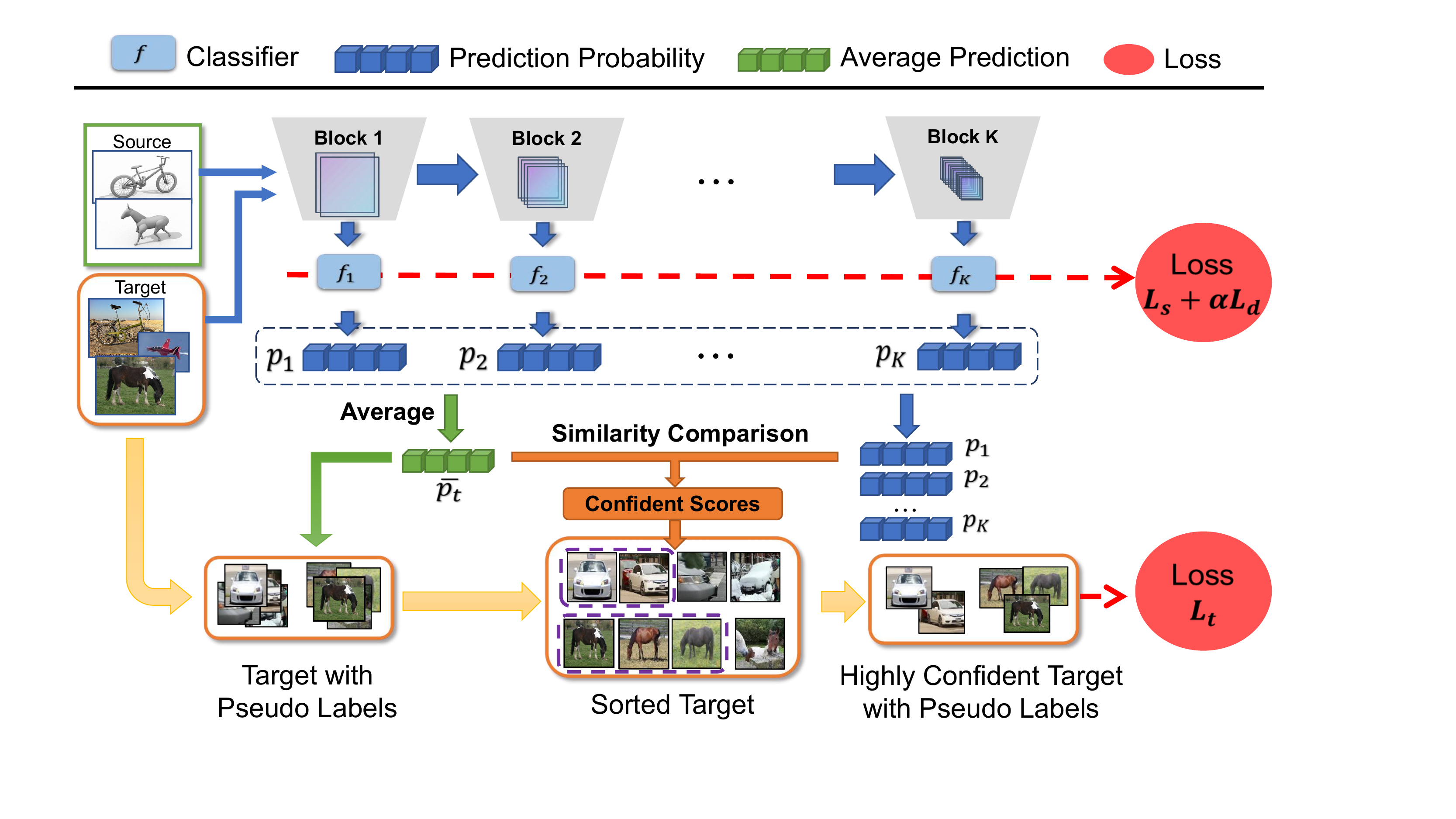}
  \caption{Illustration of the proposed dynamic domain adaptation (DDA). DDA leverages target class-balanced self-training strategy to effectively improve the transferability of all classifiers in this multi-exit architecture. Meanwhile, target inference time can be significantly accelerated by DDA.}
  \label{Fig_method}
  \end{figure}

Specifically, we denote $G = \{f_k(\cdot; \theta_k)\}_{k=1}^{K}$ as the adaptive inference model with $K$ intermediate classifiers (also called ``exit'') at the varying depth, as shown in Fig.~\ref{Fig_method}, where $f_k$ is the $k^{th}$ classifier with the corresponding parameters $\theta_k$. Notably, we expect to improve all the classifiers' transferability. The characteristics of multi-exit architecture are that the early exits can only produce coarse predictions since they only have access to coarse-level features from shallow networks, however, the last exit predicts samples more correctly due to fine-grained and global information.

Intuitively, to improve the model transferability, one can deploy domain confusion loss on each exit separately. But in such a way, the domain confusion loss will inevitably sacrifice the feature discriminability for transferability, and even deteriorate cross-domain recognition performance of some classifiers, since features in different scales have distinct transferability as shown in~\cite{transferable}. Besides, the brute-force alignment on all scale features without interaction may cause over transfer. Accordingly, a satisfactory balance between feature discriminability and transferability should be attended to. To cope with these limitations, in this work, we specially design confident target selection and self-training strategies to improve the transferability of all the classifiers without losing their recognition capacities.

\subsection{Adaptive Inference Network with Domain Confusion Learning}
Given source samples and their corresponding ground truth labels, it is straightforward to equip network with basic source classification ability. Thus, following the standard source supervised learning setting, we first use empirical risk minimization for all classifiers on source samples:
\begin{equation}\label{equ:source_loss}
  \mathcal{L}_{s}=\frac{1}{N_s}\sum_{i=1}^{N_s}\sum_{k=1}^{K}{\mathcal{E}(f_{k}(\x_i^s;\theta_{k}),{y}_i^s)},
\end{equation}
where $\mathcal{E}(\cdot, \cdot)$ is cross-entropy loss, and $f_{k}(\x_i^s;\theta_{k})$ is the probability output predicted by the $k^{th}$ classifier for $\x_i^s$.

Then, to enable all the classifiers' adaptation capacity, we can apply various domain confusion losses on each exit. Here, we take the domain adversarial loss~\cite{DANN} as an example by imposing the binary domain discriminator.
Given the source samples labeled as 0 and target samples labeled as 1, the domain discriminator can be trained with standard cross entropy loss as:
\begin{equation}\label{equ:domain_loss}
\begin{split}
  \mathcal{L}_{d}&=\frac{1}{N_s}\sum_{\x\in D_s}\sum_{k=1}^{K}{[\log D_k(F_{k}(\x;\theta_{k}))]} \\
  &+\frac{1}{N_t}\sum_{\x\in D_t}\sum_{k=1}^{K}{[\log (1-D_k(F_{k}(\x;\theta_{k})))]},
\end{split}
\end{equation}%
where $D_k(\cdot)$ is the $k^{th}$ domain discriminator, and $F_{k}(\x;\theta_{k})$ denotes the feature representations of $\x$ before the $k^{th}$ classifier. Finally, the domain-invariant representations at different scales can be achieved independently by training their corresponding feature extractors and domain discriminators adversarially. However, due to the differences of domain shift and transferability of multi-scale features as shown in \cite{transferable}, directly applying domain confusion losses may deteriorate the discriminability of each classifier.

As a consequence, it is crucial to well balance the feature transferability and discriminability in early classifiers and to effectively transfer the fine-grained and global knowledge from the later exits to the earlier predictors. To this end, in our DDA framework, we propose to select highly confident target data with their pseudo labels, and then leverage them to retrain the adaptive model for better transfer performance. Then, the speedability and recognition power of all the classifiers can be fully explored under domain shift scenarios.

\subsection{Target Confidence Score Learning Strategy}
The architecture of multi-exit network can be considered as a sequential prediction by a set of subnetworks. The earlier exits will predict samples based on coarse-level features with faster inference, and the later exits will predict the samples more correctly with much more computational cost, especially for the ``hard'' images. Hence, the prediction of the same instance may vary between classifiers. Moreover, we cannot guarantee that the last classifier would make the most correct inference for target data, as each instance could have its suitable receptive field to be recognized~\cite{zheng2020}. Thus, in this kind of multi-exit network, we propose to assign target pseudo labels via modeling certainty across all classifiers instead of using the prediction of any individual exit, which could greatly reduce possible noise in label construction for target data.

Specifically, given a target sample $\x_j^t$, we compute the average prediction $\bar{\p}_j^t=\frac{1}{K}\sum_{k=1}^{K} f_k(\x_j^t;\theta_k)$ of all classifiers as the prediction mean of the multi-exit network. We assume that the divergence between the prediction of each classifier $f_k(\x_t^j;\theta_k)$ and the average prediction $\bar{\p}_t^j$ reflects how much the classifier agrees with the result to some extent. We thus measure the agreement between them via a cosine similarity. Note that high similarities between prediction probabilities indicate high confidence of predictions.

Nevertheless, when a sample confuses all classifiers due to its difficulty, i.e., its prediction probabilities are evenly spread over classes, it is possible that the obtained confidence score would be closer to 1 as well. To avoid this, we rescale the confidence score by the max value in $\bar{\p}_t^j$, which ensures that hard examples have low confidence. Therefore, we can formulate the confidence score $v_j$ for sample $\x_t^j$ as:
\begin{equation}\label{equ:confidence_value}
  v_j=\max({\bar{\p}}_j^t)\sum_{k=1}^{K}\frac{f_k(\x_j^t;\theta_k)\cdot{\bar{\p}}_j^t}{|f_k(\x_j^t;\theta_k)||{\bar{\p}}_j^t|}.
\end{equation}

Once the confidence score set $V=\{v_j\}_{j=1}^{N_t}$ for target domain has been built, we can sort the score set by the values and select highly confident target samples with pseudo labels for the follow-up class-balanced self-training.

\subsection{Target Class-balanced Self-training Strategy}
Intuitively, the top dozens of samples would be ideal for constructing additional target self-training set. However, the confidence scores may relatively high in easy-to-transfer classes, leading to imbalanced predictions. Namely, samples with the highest confidence scores may all belong to several specific categories, which may result in model overfitting to those classes, and reduce prediction diversity.

To alleviate this issue, we propose a novel class-balanced strategy that adopts a global view for pseudo-labeled target sample selection. To be specific, for each target class $c$, we can derive the class-wise confidence score $e_c$ via accumulating the corresponding target confidence scores as:
\begin{equation}\label{equ:sum_confidence}
  e_c=\frac{1}{N_t^c}\sum_{\x_j^t\in \hat{D}_t^c}v_j,
\end{equation}
where $\hat{D}_t^c$ denotes all the target samples with their pseudo labels being $c$, and $N_t^c$ is the number of instances in $\hat{D}_t^c$.

Obviously, class-wise confidence score is varied with the transferability. For those classes with poor transferability, its corresponding class-wise confidence score is lower than others. In order to ensure that target samples under those categories can still be selected for target self-training to improve the prediction diversity, we apply a simple linear method to decide the number of selected target samples for class $c$ from our built target self-training set.
The number threshold $\lambda_c$ conditioned on class $c$ is defined as:
\begin{equation}\label{equ:class_lambda}
  \lambda_c=N_t\times\mu\frac{e_c}{\sum_{i=1}^{C}e_i},
\end{equation}
where $\mu$ is a control factor that determines the proportion of target data that will be used to construct the target self-training set $U$. Hence, to conduct target class-balanced self-training, we first select highly confident target samples according to the order of confidence score in the sorted set $V$. Given the $j^{th}$ element in $V$, we can assume its relevant target sample $\x_{(v_j)}^t$ with prediction being class $c$. At this time, if the total number of samples in target self-training subset $U_c$ of class $c$ is smaller than threshold $\lambda_c$, we take target sample $\x_{(v_j)}^t$ into $U_c$ and the size of it will be increased by 1. The process of self-training set selection can be formulated as:
\begin{equation}
 I_j^t =
  \begin{cases}
    1, &\mathrm{if}\ |U_{c}|<\lambda_c\ \mathrm{and}\ {{\hat{y}}_{(v_j)}^t}=c,\\
	0, &\mathrm{otherwise,}
  \end{cases}
\end{equation}
where $I_j^t$ is the decision function and $|U_{c}|$ is the size of target self-training subset for category $c$. We show empirically in the ablation study that this strategy works better for DDA than the classical class-balance method~\cite{CBST}.

After obtaining set $U$, the processing phase can move to self-training with highly confident and class-balanced target data.
Formally, we randomly allocate samples in $U$ to different classifiers, and we denote the targeted exit for sample $\x_j^t\in U$ as $k_j$. Then target self-training classification objective with cross-entropy loss can be formulated as:
\begin{equation}\label{equ:target_loss}
  \mathcal{L}_{t}=\frac{1}{|U|}\sum_{\x_j^t\in U}{\mathcal{E}(f_{k_j}(\x_j^t;\theta_{k_j}),{\hat{y}}_j^t)},
\end{equation}
where $|U|$ is the size of target self-training samples. Using all samples in $U$ to train each classifier may lead them to learn similar decision boundaries. So that the confidence score based on classifiers divergence will not work. This motivates us to feed each exit with different samples for self-training to ensure the diversification of their capabilities.

In summary, the overall objective function of DDA is:
\begin{equation}\label{equ:overall_loss}
  \mathcal{L}=\mathcal{L}_s+\alpha\mathcal{L}_d+\beta\mathcal{L}_t,
\end{equation}
where $\alpha$ and $\beta$ are two trade-off parameters. DDA not only leverages the proposed target class-balanced self-training to overcome the cross-domain discrepancy effectively for all the classifiers, but also speeds up the target inference significantly compared with the existing DA methods. 
The complete DDA algorithm is presented in Alg.~\ref{alg:Framwork}.

\begin{algorithm}[!htbp]
  \small
  \caption{\small Dynamic Domain Adaptation.}
  \label{alg:Framwork}
  \begin{algorithmic} [1]
    \REQUIRE
      Source domain $\{\left(\x_i^s,y_i^s\right)\}_{i=1}^{N_s}$; Target domain $\{\x_j^t\}_{j=1}^{N_t}$; Parameters $\mu$, $\alpha$ and $\beta$; Max iteration: $I$
    \ENSURE
      trained model $G=\{f_k(\cdot; \theta_k)|_{k=1}^{K}\}$
    \begin{enumerate}
        \item[\textbf{Step 1}] Adaptive Network with Domain Confusion Learning:
    \end{enumerate}
        \STATE Compute $\Lm_s$ and $\Lm_d$ with SGD optimization;
    \begin{enumerate}
        \item[\textbf{Step 2}] Target Class-balanced Self-training:
    \end{enumerate}
        \FOR{$i=1,2,\cdots,I$}
            \STATE For each $\x_j^t$, calculate $\bar{\p}_j^t$ and confidence score $v_j$;
            \STATE Sort confidence score set $V$;
            \STATE For each class $c$, calculate threshold $\lambda_c$;
            \STATE Construct target self-training set $U$;
            \STATE Randomly assign samples in $U$ to different classifiers;
            \STATE Proceed target class-balanced self-training, compute $\mathcal{L}_{s}$, $\mathcal{L}_{t}$ and $\Lm_d$ with SGD optimization;
        \ENDFOR
  \end{algorithmic}
\end{algorithm}


\begin{figure*}[ht]
  \centering
  \includegraphics[width=0.94\textwidth]{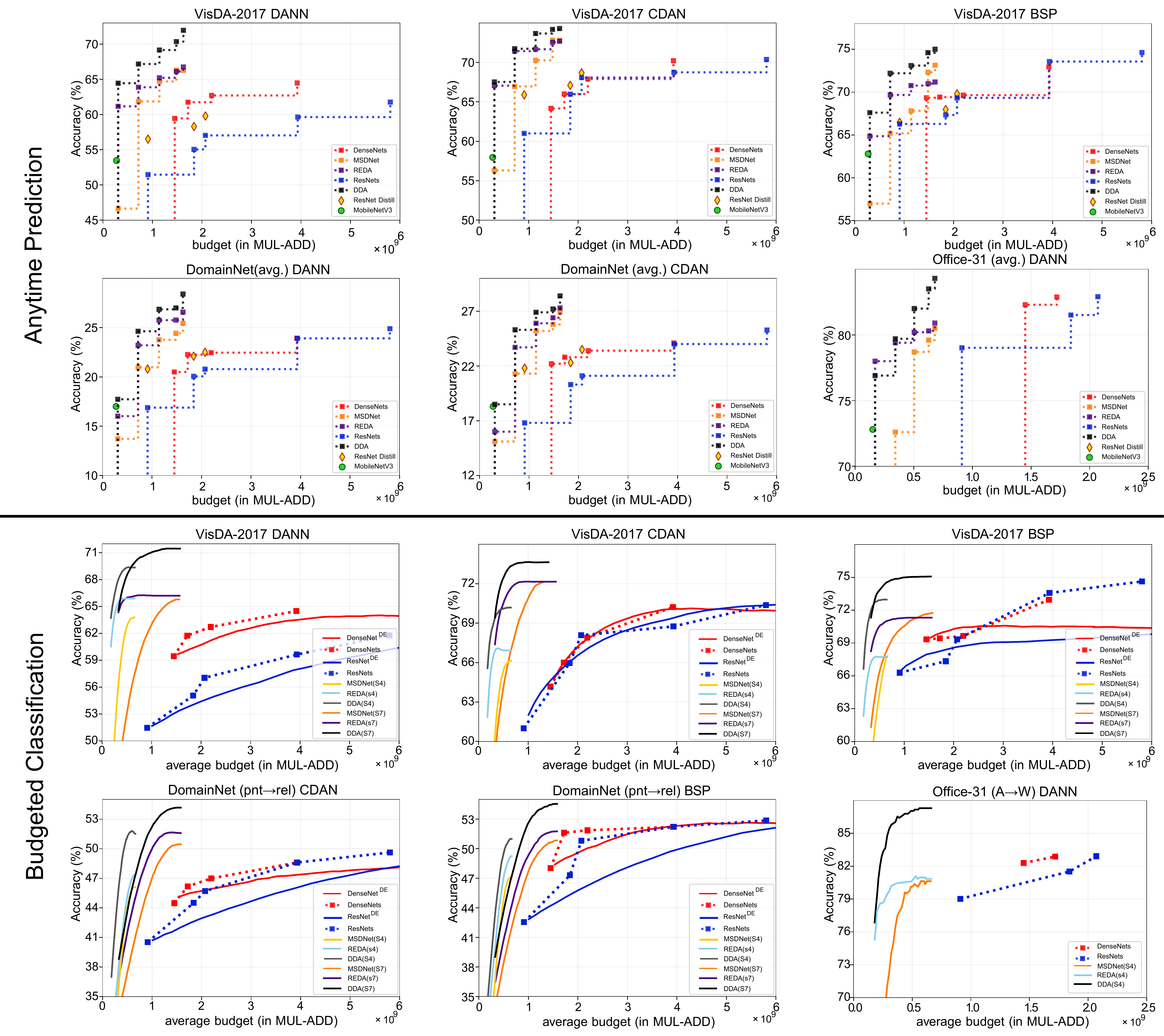}
  \caption{Anytime prediction and budgeted classification results of DA under a variant combinations of DA methods and datasets, all the networks within a subplot utilize the same DA method marked above. (best viewed in color)}
  \label{fig_main_exp}
\end{figure*}

\newcommand{\tabincell}[2]{\begin{tabular}{@{}#1@{}}#2\end{tabular}}  
\begin{table*}[htbp]
  \centering
  \caption{Accuracy (\%) on DomainNet for unsupervised domain adaption. In each sub-table, the column-wise domains are selected as the source domain and the row-wise domains are selected as the target domain.}\label{tab:addlabel}
   \resizebox{\textwidth}{!}{
  \setlength{\tabcolsep}{0.5mm}{
    \begin{tabular}{c|ccccccc|c|ccccccc||c|ccccccc|c|ccccccc}
    \toprule
    \tabincell{c}{Res50\\DANN} & {\color{blue} clp}   & {\color{blue} inf}   & {\color{blue} pnt}   & {\color{blue} qdr}   & {\color{blue} rel}   & {\color{blue} skt}   & Avg.  & \textbf{\tabincell{c}{DDA(S4)\\DANN}} & {\color{blue} clp}   & {\color{blue} inf}   & {\color{blue} pnt}   & {\color{blue} qdr}   & {\color{blue} rel}   & {\color{blue} skt}   & Avg.  & \tabincell{c}{Res152\\DANN} & {\color{blue} clp}   & {\color{blue} inf}   & {\color{blue} pnt}   & {\color{blue} qdr}   & {\color{blue} rel}   & {\color{blue} skt}   & Avg.  & \textbf{\tabincell{c}{DDA(S7)\\DANN}} & {\color{blue} clp}   & {\color{blue} inf}   & {\color{blue} pnt}   & {\color{blue} qdr}   & {\color{blue} rel}   & {\color{blue} skt}   & Avg. \\
    \hline
    {\color{blue} clp}   & -     & 13.3  & 28.4  & 9.7   & 44.3  & 29.3  & 25.0  & {\color{blue} clp}   & -     & 15.5  & 33.8  & 18.5  & 47.0  & 36.2  & 30.2  & {\color{blue} clp}   & -     & 15.6  & 33.8  & 13.1  & 50.2  & 35.6  & 29.6  & {\color{blue} clp}   & -     & 16.8  & 36.3  & 20.7  & 51.3  & 39.0  & 32.8 \\
    {\color{blue} inf}   & 19.1  & -     & 21.7  & 2.6   & 30.7  & 17.0  & 18.2  & {\color{blue} inf}   & 28.2  & -     & 26.0  & 8.4   & 38.0  & 21.1  & 24.3  & {\color{blue} inf}   & 23.8  & -     & 26.5  & 4.3   & 35.6  & 21.0  & 22.2  & {\color{blue} inf}   & 29.4  & -     & 28.1  & 10.0  & 43.8  & 23.0  & 26.9 \\
    {\color{blue} pnt}   & 29.5  & 14.2  & -     & 4.4   & 45.8  & 26.7  & 24.1  & {\color{blue} pnt}   & 37.6  & 15.9  & -     & 8.9   & 48.1  & 31.8  & 28.5  & {\color{blue} pnt}   & 35.4  & 15.8  & -     & 5.9   & 50.3  & 32.8  & 28.1  & {\color{blue} pnt}   & 40.4  & 17.2  & -     & 10.9  & 52.1  & 33.9  & 30.9 \\
    {\color{blue} qdr}   & 10.4  & 1.9   & 3.5   & -     & 7.4   & 7.1   & 6.1   & {\color{blue} qdr}   & 21.5  & 2.8   & 7.4   & -     & 15.1  & 13.0  & 11.9  & {\color{blue} qdr}   & 14.6  & 2.3   & 5.0   & -     & 12.0  & 9.1   & 8.6   & {\color{blue} qdr}   & 21.2  & 3.0   & 8.4   & -     & 18.6  & 14.1  & 13.1 \\
    {\color{blue} rel}   & 36.5  & 15.6  & 39.8  & 4.5   & -     & 26.3  & 24.6  & {\color{blue} rel}   & 43.4  & 18.1  & 41.8  & 9.4   & -     & 30.7  & 28.7  & {\color{blue} rel}   & 42.5  & 17.6  & 44.2  & 6.1   & -     & 32.5  & 28.6  & {\color{blue} rel}   & 46.0  & 18.7  & 44.9  & 11.8  & -     & 33.9  & 31.1 \\
    {\color{blue} skt}   & 37.4  & 13.2  & 33.1  & 9.1   & 41.4  & -     & 26.8  & {\color{blue} skt}   & 49.5  & 16.4  & 36.6  & 17.9  & 47.0  & -     & 33.5  & {\color{blue} skt}   & 44.8  & 16.4  & 40.3  & 12.2  & 47.3  & -     & 32.2  & {\color{blue} skt}   & 51.1  & 17.3  & 40.0  & 20.5  & 50.9  & -     & 36.0 \\
    Avg.  & 26.6  & 11.6  & 25.3  & 6.0   & 33.9  & 21.3  & 20.8  & Avg.  & 36.0  & 13.7  & 29.1  & 12.6  & 39.0  & 26.6  & \textbf{\color{red} 26.2} & Avg.  & 32.2  & 13.5  & 29.9  & 8.3   & 39.1  & 26.2  & 24.9  & Avg.  & 37.6  & 14.6  & 31.5  & 14.8  & 43.3  & 28.8  & \textbf{\color{red} 28.4} \\
    \hline
    \hline
    \tabincell{c}{Res50\\CDAN} & {\color{blue} clp}   & {\color{blue} inf}   & {\color{blue} pnt}   & {\color{blue} qdr}   & {\color{blue} rel}   & {\color{blue} skt}   & Avg.  & \textbf{\tabincell{c}{DDA(S4)\\CDAN}} & {\color{blue} clp}   & {\color{blue} inf}   & {\color{blue} pnt}   & {\color{blue} qdr}   & {\color{blue} rel}   & {\color{blue} skt}   & Avg.  & \tabincell{c}{Res50\\BSP} & {\color{blue} clp}   & {\color{blue} inf}   & {\color{blue} pnt}   & {\color{blue} qdr}   & {\color{blue} rel}   & {\color{blue} skt}   & Avg.  & \textbf{\tabincell{c}{DDA(S4)\\BSP}} & {\color{blue} clp}   & {\color{blue} inf}   & {\color{blue} pnt}   & {\color{blue} qdr}   & {\color{blue} rel}   & {\color{blue} skt}   & Avg. \\
    \hline
    {\color{blue} clp}   & -     & 13.5  & 28.3  & 9.3   & 43.8  & 30.2  & 25.0  & {\color{blue} clp}   & -     & 14.5  & 32.6  & 21.4  & 48.5  & 36.9  & 30.8  & {\color{blue} clp}   & -     & 13.8  & 28.2  & 10.1  & 44.5  & 30.8  & 25.5  & {\color{blue} clp}   & -     & 14.6  & 33.1  & 20.9  & 49.2  & 35.8  & 30.8 \\
    {\color{blue} inf}   & 18.9  & -     & 21.4  & 1.9   & 36.3  & 20.8  & 19.9  & {\color{blue} inf}   & 30.7  & -     & 28.9  & 8.1   & 42.9  & 23.2  & 26.8  & {\color{blue} inf}   & 19.6  & -     & 21.5  & 2.3   & 37.4  & 21.0  & 20.4  & {\color{blue} inf}   & 30.4  & -     & 30.0  & 8.3   & 43.3  & 23.7  & 27.2 \\
    {\color{blue} pnt}   & 29.6  & 14.4  & -     & 4.1   & 45.2  & 29.0  & 24.5  & {\color{blue} pnt}   & 38.9  & 14.8  & -     & 9.6   & 50.3  & 33.4  & 29.4  & {\color{blue} pnt}   & 32.2  & 14.7  & -     & 4.3   & 50.8  & 29.3  & 26.3  & {\color{blue} pnt}   & 38.6  & 15.0  & -     & 9.1   & 50.4  & 34.4  & 29.5 \\
    {\color{blue} qdr}   & 11.8  & 1.2   & 4.0   & -     & 9.4   & 5.7   & 6.4   & {\color{blue} qdr}   & 23.7  & 1.3   & 3.9   & -     & 16.1  & 14.4  & 11.9  & {\color{blue} qdr}   & 13.9  & 1.2   & 3.8   & -     & 10.4  & 6.6   & 7.2   & {\color{blue} qdr}   & 23.8  & 1.4   & 3.8   & -     & 17.6  & 14.3  & 12.2 \\
    {\color{blue} rel}   & 36.4  & 18.3  & 40.9  & 3.4   & -     & 26.2  & 25.0  & {\color{blue} rel}   & 44.9  & 16.8  & 43.3  & 12.1  & -     & 33.7  & 30.2  & {\color{blue} rel}   & 37.0  & 18.5  & 40.7  & 4.1   & -     & 26.8  & 25.4  & {\color{blue} rel}   & 45.0  & 18.7  & 43.1  & 12.7  & -     & 33.9  & 30.7 \\
    {\color{blue} skt}   & 38.2  & 14.7  & 33.9  & 7.0   & 36.6  & -     & 26.1  & {\color{blue} skt}   & 49.8  & 15.3  & 38.2  & 21.1  & 48.4  & -     & 34.5  & {\color{blue} skt}   & 38.8  & 14.9  & 34.4  & 8.0   & 36.8  & -     & 26.6  & {\color{blue} skt}   & 49.9  & 15.5  & 38.0  & 17.5  & 46.9  & -     & 33.6 \\
    Avg.  & 27.0  & 12.4  & 25.7  & 5.1   & 34.3  & 22.4  & 21.1  & Avg.  & 37.6  & 12.5  & 29.3  & 14.4  & 41.2  & 28.3  & \textbf{\color{red} 27.2} & Avg.  & 28.3  & 12.6  & 25.7  & 5.8   & 36.0  & 22.9  & 21.9  & Avg.  & 37.5  & 13.1  & 29.6  & 13.7  & 41.5  & 28.5  & \textbf{\color{red} 27.3} \\
    \bottomrule
    \end{tabular}%
  \label{tab:domainnet} }}%
\end{table*}%

\section{Experiment}
\subsection{Datasets and Setup}

\textbf{Office31}~\cite{Office31} is a standard dataset for DA which contains 4,652 images from 3 domains: Amazon (\textbf{A}), Webcam (\textbf{W}), Dslr (\textbf{D}). We evaluate our method on all six transfer tasks: \textbf{A}~$\rightarrow$~\textbf{W}, \textbf{W}~$\rightarrow$~\textbf{A}, $\cdots$, \textbf{W}~$\rightarrow$~\textbf{D} and \textbf{D}~$\rightarrow$\textbf~{W}.

\textbf{VisDA-2017}~\cite{VisDA2017} is a dataset for 2017 Visual Domain Adaptation Challenge~\footnote{http://ai.bu.edu/VisDA-2017/}. 
It includes over 280K images and 12 categories . Among them, the training set is synthetic images (\textbf{S}) and the validation set contains real images (\textbf{R}) collected from Microsoft COCO~\cite{Microsoft2014}. 

\textbf{DomainNet}~\cite{DomainNet} is currently the largest cross-domain benchmark. 
The whole dataset comprises $\sim$0.6 million images from 6 distinct domains: Infograph (\textbf{inf}), Quickdraw (\textbf{qdr}), Real (\textbf{rel}), Sketch (\textbf{skt}), Clipart (\textbf{clp}), Painting (\textbf{pnt}). 
Each domain has 345 categories.

All the models in our experiment are implemented using PyTorch~\cite{paszke2019pytorch}. We utilize \textbf{MSDNet(S4)} and \textbf{MSDNet(S7)} pretrained on ImageNet as the backbone network for DDA(S4) and DDA(S7) respectively. 
Note that both backbones have 5 classifier exits, but, their convolutional layers in each network block are different (i.e., 4 layers for MSDNet(S4) and 7 layers for MSDNet(S7)). To show that DDA is a general framework for most of the DA methods, we choose the classical DANN~\cite{DA_bp} and recently better-performed CDAN~\cite{CDAN}, BSP~\cite{BSP} as specifications of Eq. \eqref{equ:domain_loss}. We denote them in the form of ``DDA+\textit{method}'' (i.e., DDA+DANN). 
Moreover, the trade-off parameters $\alpha$ and $\beta$ are both selected as 1.0 using Deep Embedded Validation~\cite{DEV}, and the control factor $\mu$ is set as 80\% in all datasets without special annealing. \textbf{Code is available at} \url{https://github.com/BIT-DA/DDA}.

\subsection{Anytime Prediction}
In anytime prediction setup, the model should possess the capacity to make predictions at a randomly given time.

\textbf{Baselines.} 
We compare our DDA with the following baselines selected from all aspects: ResNet (18 to 152 layers)~\cite{resnet}, DenseNet (121 to 201 layers)~\cite{densenet}, MobileNetV3~\cite{mobilenet3}, ResNet (18, 34, 50 layers) with knowledge distillation~\cite{knowledgedistill1}, MSDNet~\cite{msdnet} and REDA~\cite{REDA}.

Among these baselines, ResNets are the most commonly used backbones for DA, and MobileNetV3 is a representative efficient network architecture. Knowledge distillation technique is implemented using ResNet101 as the teacher network and KL-divergence as the additional loss. REDA is implemented by ourselves with all the parameters consistent with the original paper. All methods are compared when applied the same DA method, i.e., ResNet18+DANN vs. REDA+DANN vs. DDA+DANN.

\textbf{Experiment results.} As presented in Fig.~\ref{fig_main_exp}(upper half), we compare DDA with different baseline nets integrate with the same DA method. 
Those results on Office31 and DomainNet are obtained by taking the average over all tasks. 
As a result, our method significantly outperforms all the baselines on these datasets. 
The last exit of DDA(S4)+DANN obtains \textbf{2.1\%} higher average-accuracy while reducing $\bf 4\times$ \textbf{FLOPs} compared to ResNet50+DANN on Office31, and DDA(S7)+CDAN outperforms ResNet152+CDAN by \textbf{4.5\%} on VisDA-2017 with \textbf{$\bf 3.6\times$ FLOPs} saving. 
On the more challenging DomainNet, DDA also achieves the best adaptability with less computational cost. Similarly, when competing against ResNets+KD and lightweight MobileNetV3, DDA reaches higher prediction accuracy than each of the counterparts using equal amount of resources. \

\textbf{Comparison to MSDNet and REDA.} When compared to MSDNet+DA which simply adds domain confusion objective to all exits, DDA achieves an average-accuracy boost of \textbf{5.5\%} (Office31-DANN), \textbf{5.6\%} (VisDA2017-CDAN) and \textbf{2.6\%} (VisDA2017-BSP). Most notably, DDA surpasses its \textit{efficient DA inference} rival REDA at every exits, especially the last one. Such result proves that the pseudo-labeled target samples selected by our two novel strategies are beneficial to \textbf{all} exits: they not only, like the knowledge distillation in REDA, improve the transferability at shallower classifiers, but also teaches the last classifier to learn better domain invariant features.

\begin{table}[htbp]
  \centering
  \caption{Accuracy (\%) on Office31 ({W} $\rightarrow$ {A}) and VisDA-2017 for unsupervised domain adaption.}
  \label{tab:office-and-VisDA}%
  \resizebox{0.48\textwidth}{!}{%
    \begin{tabular}{l|c|c|c|c|c|c}
    \toprule
    \multicolumn{1}{c|}{\multirow{2}[4]{*}{Task}} & \multicolumn{3}{c|}{ResNet50} & \multicolumn{3}{c}{\textbf{DDA(S4)}} \\
\cline{2-7}          & DANN  & CDAN  & BSP   & DANN  & CDAN  & BSP \\
  \hline
  \hline
    \multicolumn{1}{c|}{Office31 (W $\rightarrow$ A)} & 67.4  & 69.3  & 70.7  & \textbf{70.6} (3.2 $\uparrow$)  & \textbf{70.9} (1.6 $\uparrow$)  & \textbf{71.5} (0.8 $\uparrow$) \\
    \hline
    \multicolumn{1}{c|}{VisDA-2017} & 57.1  & 68.0 &  69.3 & \textbf{69.4} (12.3 $\uparrow$)  & \textbf{71.2} (3.2 $\uparrow$)  & \textbf{71.4} (2.1 $\uparrow$) \\
    \bottomrule
  \end{tabular}
  }%
\end{table}%

To better illustrate that DDA can effectively enhance the cross-domain performance on the last exit, we summarize the test accuracy of the entire 30 tasks on DomainNet in Table \ref{tab:domainnet}, and we make two pairwise comparison for networks having similar amount of parameters: ResNet50 vs. full depth of DDA(S4), ResNet152 vs. full depth of DDA(S7).
Clearly, we can see that the final exit of DDA(S4)+DANN gains an \textbf{5.4\%} average-accuracy increase over ResNet50+DANN, while the increment of DDA(S7)+DANN over ResNet152+DANN is also up to \textbf{3.5\%}. Same superiority appears in DDA+CDAN/BSP. Moreover, note that on DomainNet we conduct inductive learning and DDA(S7)+DANN prevails ResNet152+DANN on each sub-tasks. It shows that our class-balanced self-training helps to regularize a more robust decision boundary instead of merely remembers the target samples. More results shown in Table \ref{tab:office-and-VisDA} support our conclusion.

\subsection{Budgeted Classification}
In this setting, the model needs to assign resources based on the difficulty of samples to ensure that accumulated inferences are completed under a fixed computational budget.

\textbf{Baselines.} Except the baselines introduced in anytime prediction experiments, we additionally consider the ensemble of ResNet/DenseNet with dynamic inference as baselines, where an instance with high confidence will not pass through the deeper network. We follow the dynamic evaluation (DE) procedure proposed in~\cite{msdnet}, and calculate FLOPs according to the exit position of the sample. 

\textbf{Experiment results.} In Fig.~\ref{fig_main_exp}(lower half), we plot the accuracy of DDA(S4) and DDA(S7) as well as the baselines under DE. 
On Office31, the test accuracy of DDA(S4)+DANN with dynamic inference rises quickly and reaches \textbf{87.3\%} within the budget of $0.6\times10^{9}$ MUL-ADD, which is \textbf{6.5\%} higher than MSDNet+DANN. 
On VisDA-2017, both DDA(S7)+DANN and DDA(S7)+CDAN outperforms their respective counterparts substantially. In the budget range from $1\times10^{9}$ to $2\times10^{9}$ MUL-ADD, the average accuracy of DDA(S7) is \textbf{10\%} and \textbf{5\%} higher than that of ResNets and DenseNets using dynamic evaluation. 

\textbf{Comparison to REDA.} Once the model is allowed to allocate resources freely, DDA's characteristic of having large accuracy improvement in the final exit shows its advantage against REDA. With the help of a powerful full-depth classifier to deal with hard target samples, DDA obtains an overall performance enhancement under dynamic evaluation.

In summary, we conclude that DDA successfully finds the balance between adaptation performance and computational cost under this budgeted classification setting.


\subsection{Insight Analysis}
In this subsection, we carry out experiments to analyze the effectiveness of DDA and further investigate the influence of each component. All the analytical experiments are conducted on \textbf{VisDA-2017} using \textbf{MSDNet(S4)} as backbone and \textbf{DANN} as the adversarial objective.

\begin{figure}[htbp]
  \centering
  \includegraphics[width=0.48\textwidth]{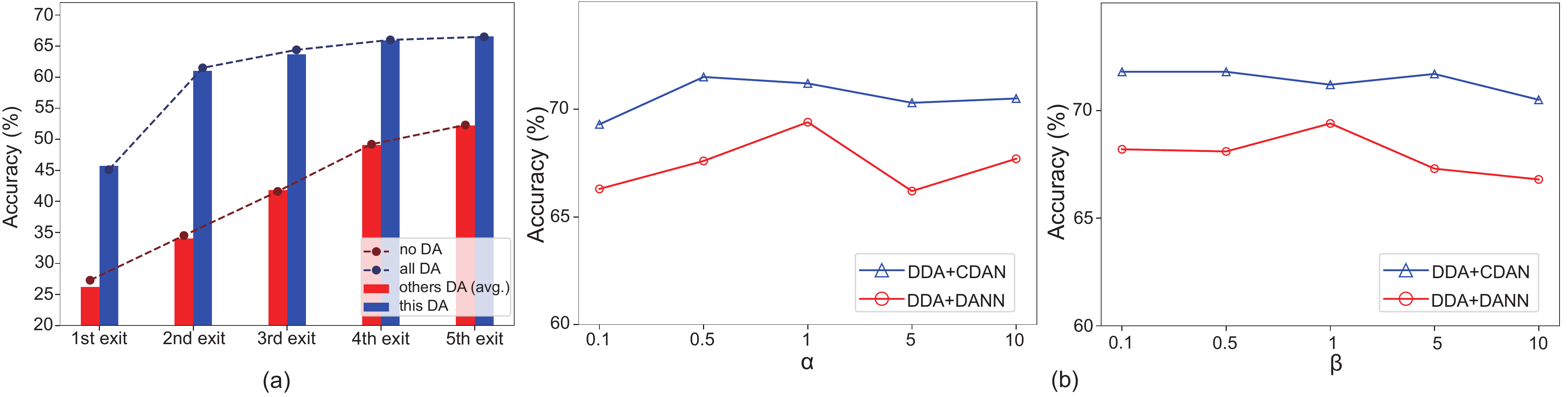}
  \caption{(a) Transferability analysis by conducting exclusive employment of DA at each exit. (b) Sensitivity analysis of $\alpha$ and $\beta$.
  }\label{fig_analysis_a}
\end{figure}
\vspace{-5pt}

\textbf{Transferability Analysis.} In our proposed framework, we deploy domain confusion loss on each exit of the backbone to improve the transferability.
Therefore, a natural question arises: why adopting DA method at all exits? Will adopting DA method at the early stage of the network influence the feature learning in latter stage?
To investigate the impact on the performance by applying domain confusion objective on classifier exits of DDA, we conduct exclusive employment of DA, one exit at a time. (We only add DA loss to one of the five exits.)
Meanwhile, we consider other variants including no-classifier DA (no DA) and all-classifier DA (all DA) as reference. 
From the result shown in Fig.~\ref{fig_analysis_a}(a), we find out that adopting DA on a particular exit only improves the accuracy of that one while having nearly no effect on the accuracy of others.
Specifically, when adopting exclusive DA on the $n^{th}$ exit, the accuracy of it (this DA) reaches the all-DA level, while other four exits remain an accuracy at the no-DA level. Thus, we can verify the validation of multi-classifier domain adaptation.

\begin{table}[htbp]
  \centering
  \caption{Confidence Score vs. Handcrafted Threshold.}
  \resizebox{0.48\textwidth}{!}{  
  \begin{tabular}{c|c|ccccc}
    \toprule
    Methods & Threshold & 1st exit  & 2nd exit  & 3rd exit  & 4th exit & 5th exit \\
    \hline
    \hline
    Handcrafted & $>0.6$   & 56.1 & 60.3  & 61.8  & 62.7  & 63.1 \\
    Handcrafted & $>0.7$   & 54.8  & 60.7  & 63.5  & 63.9  & 63.9 \\
    Handcrafted & $>0.8$   & 52.1  & 59.2  & 61.9  & 63.3  & 63.3 \\
    Handcrafted & $>0.9$   & 48.1  & 57.4  & 61.2  & 62.6  & 63.1 \\
    \hline
    Confidence (Ours)  & --     & \textbf{64.1}  & \textbf{65.2} & \textbf{67.5} & \textbf{68.4} & \textbf{69.4} \\
    \bottomrule
    \end{tabular}%
  \label{tab:threshold} }%
\end{table}%
\vspace{-5pt}

\textbf{Sensitivity Analysis.} To show that DDA is robust to hyperparameter choices, we vary the values of $\alpha$ and $\beta$ on VisDA-2017 and plot the results in Fig.~\ref{fig_analysis_a}(b). We can see that DDA achieves stable accuracies in spite of the parameter change. Actually, we find $\alpha=1$,$\beta=1$ achieves satisfying results on all datasets with no need for special annealing.

\begin{figure}[htbp]
  \centering
  \includegraphics[width=0.48\textwidth]{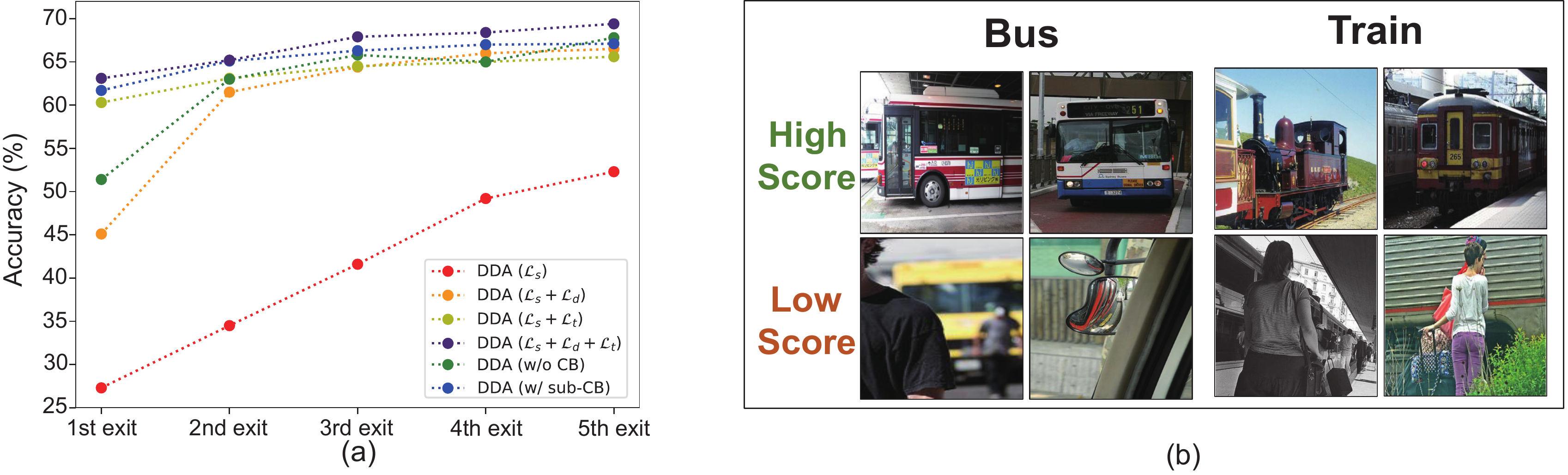}
  \caption{(a) Ablation studies of different DDA components. (b) Visualization of target samples with different confidence scores.
  }\label{fig_analysis_b}
\end{figure}

\textbf{Ablation Studies.} To discuss the contribution of different parts in DDA, we firstly study different loss combinations of it.
Moreover, to validate our proposed class-balance strategy, we (1) remove the class balance (CB) and (2) substitute CB with the classical class-balance method~\cite{CBST} and denote them as \textbf{DDA (w/o CB)} and \textbf{DDA (w/ sub-CB)}.
The result is reported in Fig.~\ref{fig_analysis_b}(a).

\vspace{-5pt}

\textbf{Confidence Score vs. Handcrafted Threshold.}
Handcrafted threshold strategy, where samples with the predicted probability higher than a specified threshold are selected, is a classical method for single-exit self-training.
Here we set the threshold as $\{0.6-0.9\}$ and compare them with our confidence score learning strategy.
The results in Table \ref{tab:threshold} show that our DDA with confidence score generation achieves superior performance than with handcrafted threshold in the case of multi-exit self-training.

\textbf{Visualization.} To illustrate that DDA is able to generate the pseudo labeled training set with less noise, we take some samples of high \& low confidence scores from two categories and show them in Fig.~\ref{fig_analysis_b}(b). 
We see that the samples in class-balanced self-training set have distinctive features and preferably single object. 
Such result verifies that DDA can effectively select confident samples for self-training.

\section{Conclusion}
In this paper, we propose a Dynamic Domain Adaptation (DDA) framework, which aims to solve the problem of efficient inference in the context of domain adaptation (DA).
Our method introduces multi-exist adaptive architecture into DA and applies domain confusion objectives.
We also design a novel self-training scheme based on confidence score strategy and class-balanced self-training strategy across classifiers.
To preserve the diversity in network predictions among exits, we randomly assign the pseudo-labeled target samples to different exits for training.
Extensive experiments on three benchmarks demonstrate that DDA substantially outperforms baseline methods as well as previous efficient DA inference models in both anytime and budgeted predictions. This proves that DDA provides a faster and better inference solution within DA.

{\small
\bibliographystyle{ieee_fullname}
\bibliography{Reference_CVPR2021}
}


\end{document}